\documentclass[conference]{IEEEtran}
\IEEEoverridecommandlockouts
\usepackage{amsmath,amssymb,amsfonts}
\usepackage{graphicx}
\usepackage{booktabs}
\usepackage{array}
\usepackage{hyperref}
\usepackage{bm}
\usepackage{mathtools}
\usepackage{enumitem}
\usepackage{caption}
\usepackage{dsfont}
\usepackage[export]{adjustbox}

\hypersetup{hidelinks}

\title{Group Evidence Matters: Tiling-based Semantic Gating for Dense Object Detection}

\author{
\IEEEauthorblockN{Yilun Xiao}
\IEEEauthorblockA{Department of Computer Science\\
University of Toronto\\
Toronto, Canada\\
\texttt{yilun.xiao@mail.utoronto.ca}}
}

\begin{document}
\maketitle

\begin{abstract}
Dense small objects in UAV imagery are often missed due to long-range viewpoints, occlusion, and clutter. This paper presents a detector-agnostic post-processing framework that converts overlap-induced redundancy into group evidence. Overlapping tiling first recovers low-confidence candidates. A Spatial Gate (DBSCAN on box centroids) and a Semantic Gate (DBSCAN on ResNet-18 embeddings) then validate group evidence. Validated groups receive controlled confidence reweighting before class-balanced NMS (CB\mbox{-}NMS) fusion. Experiments on VisDrone show a recall increase from 0.685 to 0.778 (+0.093) and a corresponding precision drop from 0.801 to 0.595, yielding F1 = 0.669. Post-processing latency averages 0.095 s per image. These results indicate recall-first, precision-trade-off behavior that benefits recall-sensitive applications such as far-field counting and monitoring. Ablation confirms that tiling exposes missed objects, spatial clustering stabilizes geometry, semantic clustering enforces appearance coherence, and reweighting provides calibrated integration with the baseline. The framework requires no retraining and integrates with modern detectors. Future work will reduce semantic gating cost and extend the approach with temporal cues.
\end{abstract}


\section{Introduction}
In recent years, the rapid proliferation of unmanned aerial vehicle (UAV) technology has made aerial vision systems indispensable in urban governance, disaster response, and public security \,[1], [2], [3]. These applications require robust detection of small and densely distributed objects under complex conditions. Long-distance imaging, severe occlusions, and dynamic backgrounds reduce the signal-to-noise ratio and increase missed detections. As a result, achieving both high recall and high precision becomes difficult \,[4], [5]. Most research on dense object detection has centered on model design. Feature Pyramid Networks (FPN) enhance multi-scale feature representation \,[6], while attention mechanisms help focus on salient regions \,[7], [8]. Such solutions, however, demand heavy retraining and incur high deployment costs. Post-processing strategies remain less explored: traditional Non-Maximum Suppression (NMS) and its variants (e.g., learning-based/improved NMS [9], Soft-NMS [10]) still rely mainly on geometric overlap. Furthermore, improvements in training objectives, such as IoU-variant losses (e.g., DIoU/CIoU) [11], likewise do not recover low-confidence objects filtered early during inference. This limitation is particularly severe in UAV imagery, where small and occluded objects are easily suppressed \,[4], [5].

Inspired by human vision, which leverages contextual cues to interpret ambiguous objects, this paper proposes a novel post-processing paradigm that transforms redundancy into decision signals. Instead of discarding overlapping detections from tiled inference, this paper’s method treats them as \emph{Group Evidence}: clusters of consistent low-confidence boxes suggest true objects, while isolated boxes are likely false positives. This approach raises a fundamental question: can redundancy-driven recovery strategies jointly achieve high recall and high precision, or are they inherently constrained by a precision--recall trade-off? The experiments confirm the latter: aggressive recovery increases recall (e.g., from 0.685 to 0.778) but reduces precision (from 0.801 to 0.595). Single-frame post-processing thus faces an intrinsic trade-off between recall and precision, requiring task-specific balancing. The main contributions of this paper can be summarized as follows:
\begin{enumerate}[leftmargin=1.5em]
\item \textbf{Group-Evidence Framework:} This paper \textbf{proposes} a detector-agnostic post-processing pipeline that leverages collective detection patterns from overlapping tiles to recover missed objects in dense scenes---without retraining the base detector.
\item \textbf{Dual-Clustering Gating Mechanism:} This paper \textbf{designs} a two-stage gating process: spatial gating (DBSCAN on centroids) filters geometrically concentrated candidates, and semantic gating (DBSCAN on normalized ResNet-18 embeddings, cosine distance) preserves visually consistent groups.
\item \textbf{Comprehensive Empirics:} This paper \textbf{conducts} a two-stage parameter study and ablations on VisDrone2019-DET, revealing a precision--recall trade-off inherent to single-frame post-processing and providing practical guidance for recall-first deployments.
\end{enumerate}

\section{Related Work}
Dense object detection in aerial imagery has attracted significant attention due to its broad practical importance. Several methods have been proposed to address the challenges of small and densely distributed objects. For instance, ClusDet introduces a clustered detection framework that unifies object clustering and detection in an end-to-end manner, leveraging cluster proposals and scale estimation to improve efficiency and accuracy in aerial scenes \,[12]. More recently, CrowdDiff formulates crowd density estimation as a generative task using diffusion models, producing high-fidelity density maps and leveraging the stochastic nature of diffusion to improve robustness in dense scenes \,[13]. In addition, large-scale aerial benchmarks such as VisDrone \,[4], DOTA \,[14], and xView \,[15] have further stimulated research on dense and small-object detection in UAV imagery. While these approaches significantly advance the state of the art, they require substantial architectural modifications and retraining, which limits their general applicability in practical deployments.

The slicing approach, also referred to as the sliding window technique, has been widely adopted for the analysis of high-resolution images. The fundamental principle is to partition the image into overlapping subregions, which reduces computational cost while enhancing the visibility of a small-scale objects. Early studies primarily utilized slicing as a preprocessing strategy, as seen in region proposal--based detectors \,[16]. More recent research, however, has further systematized this idea. For instance, the SAHI framework introduces slice-assisted inference, where multiple detections are performed within overlapping regions and subsequently merged during the post-processing stage, significantly improving the recall rate of small and densely distributed objects \,[17]. In parallel, methods such as YOLOv5-tiling incorporate the slicing operation directly into the training pipeline, enabling the model to adapt to locally high-resolution inputs during the learning phase \,[18]. Despite these advances, most existing approaches regard the redundant detections generated by slicing as noise, emphasizing techniques such as merging or Non-Maximum Suppression (NMS) and its variants (e.g., learning-based/improved NMS [9]) to mitigate their adverse effects. In contrast, this work presents a distinct perspective: slicing redundancy is not merely extraneous but can serve as a valuable source of group evidence. Low-confidence candidate boxes that would otherwise be discarded may provide reliable detection evidence when they exhibit spatial and semantic consistency with other candidates. This study aims to investigate strategies for leveraging such redundant information and to examine the practical limits of its effectiveness.

Beyond standard NMS, a variety of enhanced post-processing techniques have been proposed to alleviate the recall deficiency in dense scenes. Soft-NMS reduces the scores of overlapping candidates instead of discarding them outright, thereby retaining some potentially correct detections \,[10]. Adaptive NMS dynamically adjusts the IoU threshold according to the local object density, making it better suited to both dense and sparse scenarios \,[19]. In addition, methods that diversify proposals can increase coverage in crowded scenes \,[20]. In contrast to the above methods, the framework proposed in this work requires neither modifications to the detector architecture nor retraining. Instead, it operates entirely at the inference stage: tiling inference is first employed to generate additional candidates, which are then subjected to dual filtering based on spatial clustering and semantic consistency verification. Finally, a quality-based scoring mechanism is applied to reweigh the confidence scores.

\begin{figure*}[!t]
\centering
\includegraphics[width=\textwidth,keepaspectratio]{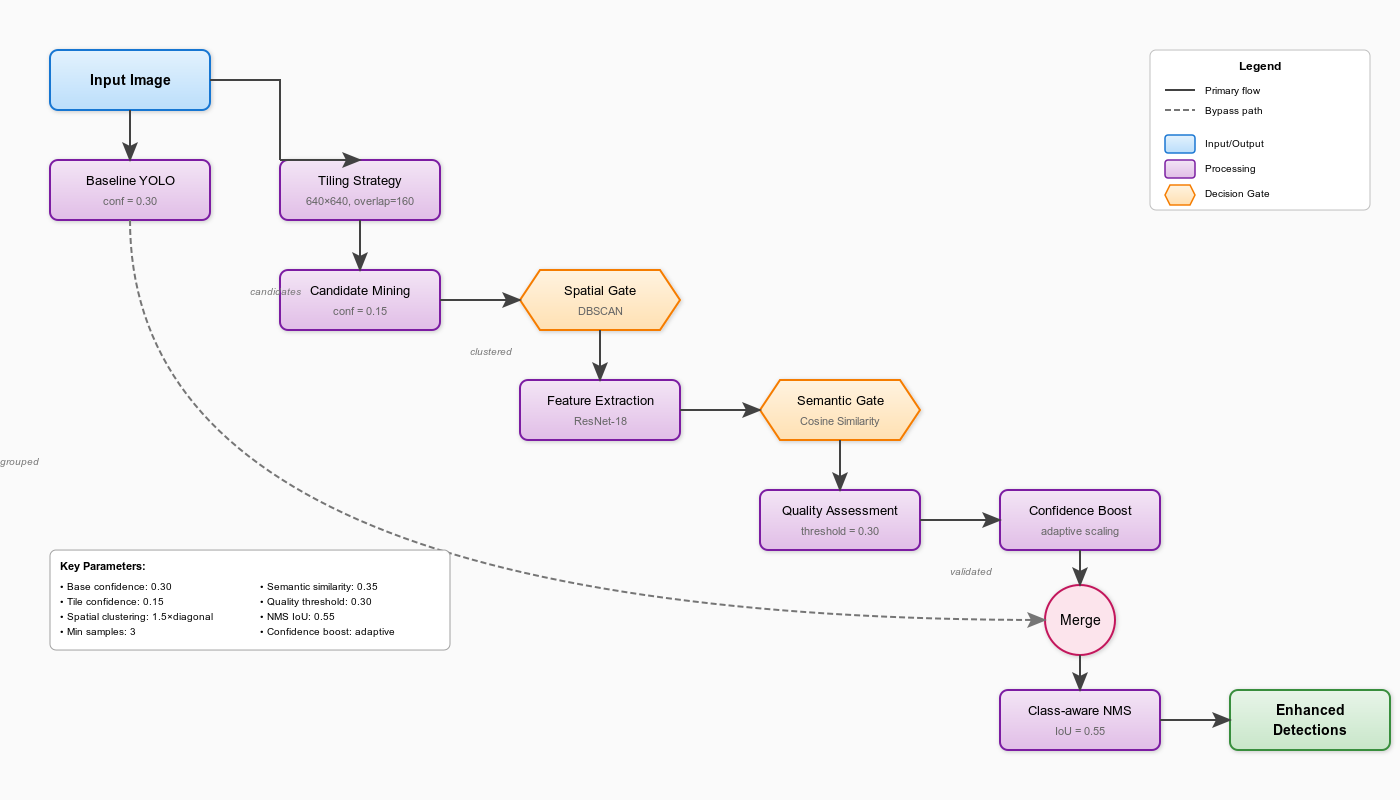}
\caption{Proposed post-processing pipeline. The baseline path preserves high-confidence detections ($\tau_{\text{base}}{=}0.30$); the enhancement path mines low-confidence candidates ($\tau_{\text{tile}}{=}0.15$) via tiling ($640{\times}640$, $160$-px overlap). Candidates pass through a Spatial Gate (DBSCAN) and a Semantic Gate (DBSCAN on normalized ResNet-18 embeddings, cosine distance). Only groups whose quality score exceeds $0.30$ receive quality-aware confidence reweighting. Finally, class-balanced NMS (CB\mbox{-}NMS [26]; $\mathrm{IoU}{=}0.55$) fuses baseline and validated candidates.}
\label{fig:pipeline}
\end{figure*}

\section{Method}
We augment baseline predictions with redundancy-driven recovery and gating. As illustrated in Fig.\,\ref{fig:pipeline}, the pipeline comprises four components: (1) Tile-based Candidate Generation; (2) Spatial Gate to consolidate spatially concentrated candidates; (3) Semantic Gate to retain appearance-consistent groups; and (4) Quality-aware Reweighting to rescore validated candidates, followed by class-balanced NMS for final fusion.

\subsection{Tile-based Candidate Generation}
Given an input image $I \in \mathbb{R}^{H \times W}$, the image is partitioned into a set of overlapping square tiles of size $T \times T$ with an overlap of $O$ pixels, a common strategy in high-resolution image analysis \,[16], [17]. Let $S = T-O$ be the stride. The tile set is formally defined by the grid points:
\begin{equation}
\begin{split}
\mathcal{T} = \{\, &I[x:x{+}T,\; y:y{+}T] \;\mid\; \\
&x \in \{0, S, 2S, \dots\},\; y \in \{0, S, 2S, \dots\}; \\
&x \le W{-}T,\; y \le H{-}T \,\}.
\end{split}
\end{equation}
Each tile $t \in \mathcal{T}$ is independently processed by the baseline detector. To balance recall and precision, a confidence relaxation strategy is adopted: detections on the full image use a conservative threshold to ensure reliability, whereas tile-level detections employ a lower threshold to aggressively capture potential true positives ($\tau_{\text{tile}}{=}0.15$, $\tau_{\text{base}}{=}0.30$ in this experiment). The union of tile-level results forms the candidate pool
\begin{equation}
\mathcal{D}_{\text{cand}} = \bigcup_{t \in \mathcal{T}} \mathrm{detector}(t;\, \tau_{\text{tile}}).
\end{equation}
To reduce the computational burden in subsequent clustering, we perform a fast duplicate removal. For each candidate detection $d{=}\{b^{xy}, l\}$, where $b^{xy}{=}(x_1,y_1,x_2,y_2)$ denotes the bounding box coordinates and $l$ the predicted class label, we compute a simple hash function
\begin{equation}
h(d) = \mathrm{hash}(b^{xy}, l),
\end{equation}
and eliminate detections with identical hash values. This step effectively removes redundant bounding boxes arising from overlapping tiles while preserving diversity in $\mathcal{D}_{\text{cand}}$.

\subsection{Spatial--Semantic Dual Gate}
DBSCAN \,[21] is first applied to identify groups of candidate boxes in the tile-generated pool $\mathcal{D}_{\text{cand}}$ that are spatially concentrated. Let $b^{xy}_i{=}(x_1,y_1,x_2,y_2)$ and the centroid be $\mathbf{p}_i{=}\big(\tfrac{x_1{+}x_2}{2},\, \tfrac{y_1{+}y_2}{2}\big)$. Spatial clusters are obtained as
\begin{equation}
\mathcal{C}_{\text{spatial}} = \mathrm{DBSCAN}\!\left(\{\mathbf{p}_i\}, \varepsilon_{\text{spatial}}, \text{min\_samples}{=}3\right),
\end{equation}
where $\varepsilon_{\text{spatial}}$ adapts to the global object scale by tying it to the average diagonal length of candidates:
\begin{equation}
\varepsilon_{\text{spatial}} = 1.5\,\overline{\mathrm{diag}(b_i)},\quad
\mathrm{diag}(b_i)=\sqrt{(x_2-x_1)^2+(y_2-y_1)^2}.
\end{equation}
We retain only clusters with at least three members and discard DBSCAN noise points (label $-1$). This gate yields a set of spatially validated groups $\{S_j\}$.

Then, for each spatial cluster $S_j$, we evaluate appearance consistency via feature-space clustering. Each candidate’s ROI is cropped from the original image and encoded by an ImageNet-pretrained ResNet-18 \,[22], producing a 512-D descriptor \,[23], [24]
\begin{equation}
\mathbf{f}_i = \mathrm{ResNet18}(\mathrm{ROI}_i) \in \mathbb{R}^{512},\qquad \hat{\mathbf{f}}_i = \frac{\mathbf{f}_i}{\|\mathbf{f}_i\|_2}.
\end{equation}
DBSCAN is then applied again in the normalized feature space using cosine distance. Because the descriptors are unit-normalized, cosine distance simplifies to
\begin{equation}
d_{\cos}(\hat{\mathbf{f}}_a,\hat{\mathbf{f}}_b) = 1 - \hat{\mathbf{f}}_a^\top \hat{\mathbf{f}}_b,
\end{equation}
with $\varepsilon_{\text{semantic}}{=}0.35$ and $\text{min\_samples}{=}3$. Let $\mathcal{C}_{\text{sem}}(S_j)$ denote the semantic sub-clusters within $S_j$ after removing noise. Only candidates belonging to these appearance-consistent sub-clusters proceed to the subsequent quality scoring and fusion stage.

\subsection{Cluster Quality Assessment and Quality-aware Reweighting}
Spatial--semantic clustering produces candidate groups $\{C^\ast\}$ with varying reliability. To systematically evaluate their credibility, a cluster quality score $Q(C^\ast)$ is introduced which integrates both confidence reliability and semantic coherence:
\begin{equation}
Q(C^\ast) = \omega_1\, Q_{\text{score}}(C^\ast) + \omega_2\, Q_{\text{consistency}}(C^\ast),
\end{equation}
where
\begin{equation}
Q_{\text{score}}(C^\ast) = \frac{1}{|C^\ast|}\sum_{s_i \in C^\ast} s_i
\end{equation}
denotes the mean confidence across cluster members ($s_i$ being the score of candidate $i$), and
\begin{equation}
Q_{\text{consistency}}(C^\ast) = \frac{\max_{l'} \sum_{i \in C^\ast} \mathds{1}(l_i = l')}{|C^\ast|}
\end{equation}
measures intra-cluster label consistency (the proportion of the majority class $l'$). Here, $\omega_1{=}0.7$ and $\omega_2{=}0.3$ are empirically chosen weights. Only clusters with $Q(C^\ast){>}0.3$ are retained as high-quality groups.

For each candidate $b_j$ in a retained cluster $C_i^\ast$, its confidence score is adjusted according to cluster reliability and scale:
\begin{equation}
s'_j = s_j \cdot \Big(1 + \beta \cdot \log\big(1 + |C_i^\ast|\big) \cdot Q(C_i^\ast)\Big).
\end{equation}
Here $|C_i^\ast|$ is the cluster size, and $\beta{=}0.1$ controls the enhancement strength. The logarithmic factor ensures smooth and bounded amplification, preventing very large clusters from dominating the final output. This principled reweighting draws inspiration from prior works on group consistency and context-driven re-scoring \,[25], while maintaining stability in dense aerial scenes.

\subsection{Final Fusion and Output}
In the final stage, the high-confidence baseline detections $\mathcal{D}_{\text{base}}$ are merged with the set of validated and confidence-boosted candidates, $\mathcal{D}'_{\text{validated}}$. To reconcile overlapping predictions while preserving dense targets, a class-balanced non-maximum suppression (CB\mbox{-}NMS) \,[26] is applied:
\begin{equation}
\mathcal{D}_{\text{final}} = \mathrm{CB\mbox{-}NMS}\!\left(\mathcal{D}_{\text{base}} \cup \mathcal{D}'_{\text{validated}},\, \mathrm{IoU}\right),
\end{equation}
with an IoU threshold of $\mathrm{IoU}{=}0.55$. This threshold achieves a balance between eliminating redundancies and maintaining recall in highly crowded scenes. By integrating baseline reliability, tile-based recovery, dual clustering gating, and quality-aware reweighting into a unified pipeline, the framework provides a plug-and-play post-processing module that can consistently enhance the detection output of existing object detectors, particularly in challenging dense-object scenarios.

\section{Experiment}
\begin{figure*}[!t]
\centering
\includegraphics[width=\textwidth,keepaspectratio]{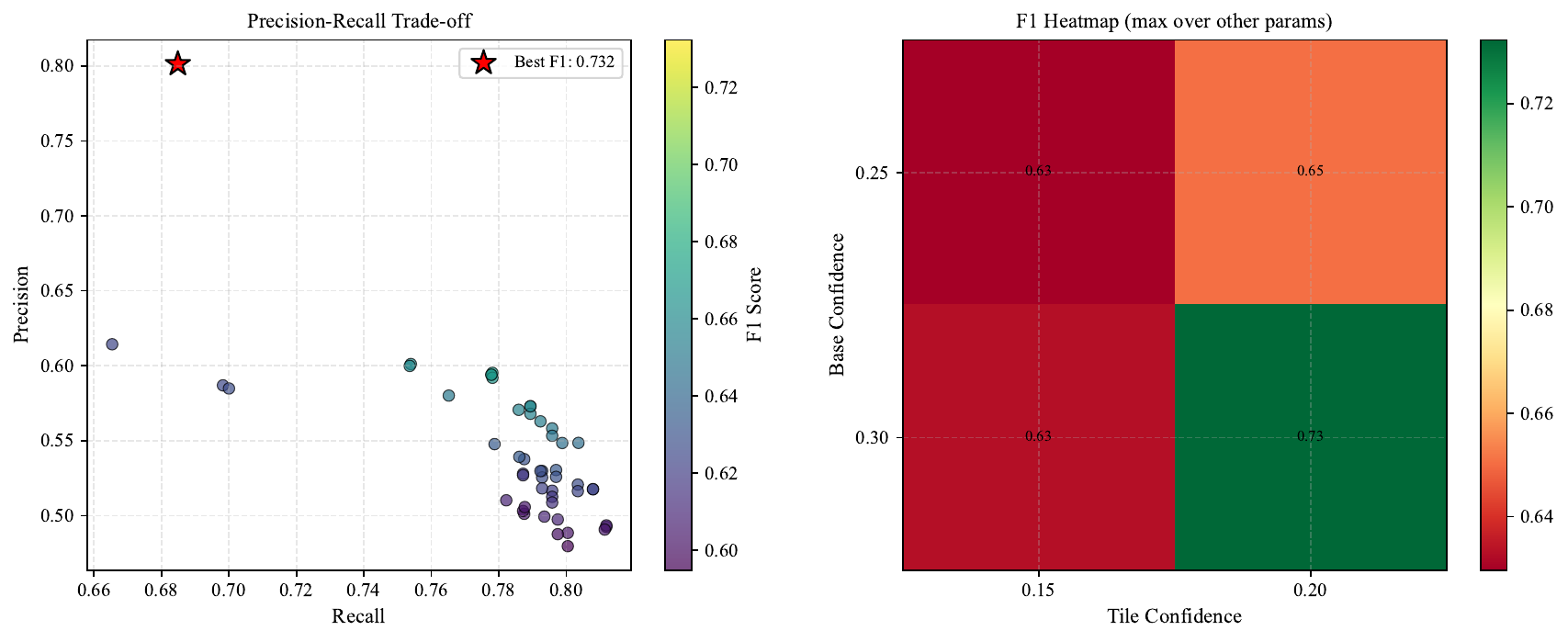}
\caption{Precision--Recall trade-off across candidate configurations. Each dot represents one configuration, with F1 encoded by both color and marker size. Stage-A (grid search) clusters high-recall/low-precision variants, while Stage-B (focused random search) pushes the Pareto frontier outward to yield more balanced trade-offs.}
\label{fig:pr_tradeoff}
\end{figure*}

\begin{figure}[t]
\centering
\includegraphics[width=\linewidth]{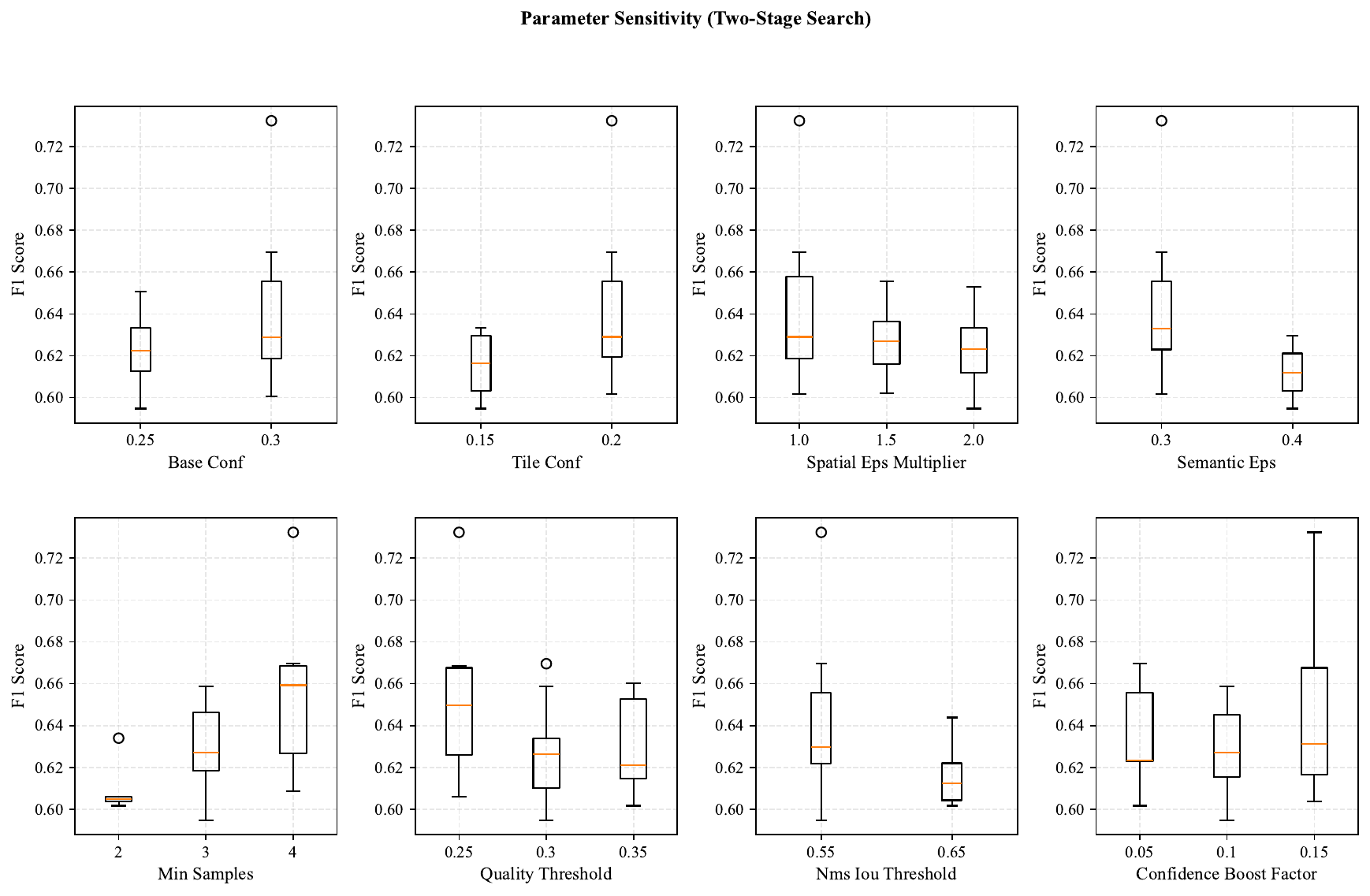}
\caption{Parameter sensitivity analysis across eight hyperparameters. Boxplots show F1-score distributions for different values. $\text{base\_conf}$ and $\text{tile\_conf}$ have the most dominant effect, whereas $\text{nms\_iou}$ and $\text{confidence\_boost\_factor}$ act as stabilizers, confirming the staged search design.}
\label{fig:param_sens}
\end{figure}

This section details the dataset, evaluation metrics, and implementation specifics, followed by a two-stage parameter study (Sec.\,\ref{sec:param_study}), an ablation study (Sec.\,\ref{sec:ablation}), a runtime analysis (Sec.\,\ref{sec:runtime}), and qualitative results (Sec.\,\ref{sec:qualitative}). Key numerical results are reported in Table~\ref{tab:top_configs} (top configurations) and Table~\ref{tab:ablation} (ablation), while Figs.\,\ref{fig:pr_tradeoff}--\ref{fig:qualitative_gallery} visualize the search distribution, ablation trajectory, runtime behavior, and qualitative comparisons.

\subsection{Dataset and Evaluation Metrics}
All experiments were conducted on the VisDrone benchmark \,[4], a large-scale UAV dataset characterized by dense aerial scenes with numerous small objects and severe occlusions. Ground-truth bounding boxes and class labels were obtained from the official annotations. Evaluation followed the standard object detection protocol: predicted boxes are matched to ground truth using an Intersection over Union (IoU) threshold of 0.5. For each image, Precision, Recall, and F1-score are computed, and the arithmetic mean across all images is reported.

\begin{table}[t]
\centering
\caption{Experimental Setup and Default Hyperparameters}
\label{tab:setup}
\begin{tabular}{@{}p{0.45\linewidth} p{0.45\linewidth}@{}}
\toprule
\textbf{Component} & \textbf{Setting} \\
\midrule
Tile size / overlap & $640 \times 640$ with $160$-pixel overlap \\
Confidence thresholds & $\tau_{\text{base}}=0.30,\;\tau_{\text{tile}}=0.15$ \\
DBSCAN (spatial) & $\varepsilon_{\text{spatial}} = 1.5 \times \text{avg diag},\; \text{min\_samples}=3$ \\
DBSCAN (semantic) & $\varepsilon_{\text{semantic}} = 0.35,\; \text{min\_samples}=3$ \\
Feature extractor & ResNet-18 \,[22] \\
Confidence reweighting & $\beta=0.1,\; Q(C^\ast)>0.3$ \\
Final NMS & Class-balanced NMS (CB\mbox{-}NMS) \,[26], IoU $=0.55$ \\
\bottomrule
\end{tabular}
\end{table}

\begin{table}[t]
\centering
\caption{Top-ranked configurations from the two-stage search. Each row reports mean Precision, Recall, F1, and post-processing time per image. While the baseline attains the highest F1, variants such as B18 and B14 provide competitive recall-first trade-offs.}
\label{tab:top_configs}
\begin{tabular}{@{}lcccc@{}}
\toprule
\textbf{Configuration} & \textbf{Precision} & \textbf{Recall} & \textbf{F1-score} & \textbf{Time (s/img)} \\
\midrule
Baseline & 0.801 & 0.685 & 0.732 & 0.001 \\
B18 & 0.595 & 0.778 & 0.670 & 0.107 \\
B10 & 0.594 & 0.778 & 0.669 & 0.278 \\
+Reweighting & 0.595 & 0.778 & 0.669 & 0.095 \\
+Semantic & 0.592 & 0.778 & 0.667 & 0.102 \\
B01 & 0.601 & 0.754 & 0.660 & 0.262 \\
B14 & 0.600 & 0.754 & 0.659 & 0.091 \\
A19 & 0.573 & 0.789 & 0.659 & 0.335 \\
B03 & 0.573 & 0.789 & 0.659 & 0.324 \\
B15 & 0.571 & 0.786 & 0.656 & 0.098 \\
A21 & 0.568 & 0.789 & 0.656 & 0.315 \\
A23 & 0.563 & 0.792 & 0.653 & 0.254 \\
\bottomrule
\end{tabular}
\end{table}

\begin{table}[t]
\centering
\caption{Ablation across progressively enabled modules. Precision, Recall, F1, and post-processing time are reported at each step, illustrating the roles of Tiling $\to$ Spatial Gate $\to$ Semantic Gate $\to$ Quality-aware Reweighting.}
\label{tab:ablation}
\begin{tabular}{@{}lcccc@{}}
\toprule
\textbf{Configuration} & \textbf{Precision} & \textbf{Recall} & \textbf{F1-score} & \textbf{Time (s/img)} \\
\midrule
Baseline   & 0.801 & 0.685 & 0.732 & 0.001 \\
+Tiling    & 0.499 & 0.793 & 0.608 & 0.002 \\
+Spatial   & 0.530 & 0.792 & 0.628 & 0.003 \\
+Semantic  & 0.592 & 0.778 & 0.667 & 0.102 \\
+Reweighting     & 0.595 & 0.778 & 0.669 & 0.095 \\
\bottomrule
\end{tabular}
\end{table}

Additionally, the average per-image processing time is recorded to assess computational efficiency. This evaluation protocol is particularly suited to dense small-object detection, where reducing missed detections (i.e., maximizing recall) is critical, while still maintaining a competitive balance with precision as reflected in the overall F1-score \,[4], [14], [15].

\subsection{Implementation and Environment}
This experiment is built on Ultralytics YOLO \,[28] as the baseline detector. All experiments are run on an NVIDIA GeForce RTX 4060 with a fixed random seed (2025). Unless otherwise stated, Table~\ref{tab:setup} lists the default post-processing hyperparameters, which are used throughout. The design is detector-agnostic and requires no retraining. To ensure fair comparisons and avoid redundant computation, we cache baseline and tiling detections. Unless explicitly noted, all reported latency refers to post-processing only (excluding base detector inference).

\subsection{Two-Stage Parameter Study Protocol}\label{sec:param_study}
To efficiently explore the parameter space while avoiding prohibitive computational costs, the experiment employed a two-stage search strategy. A coarse grid search (Stage-A) was first conducted on a 12-image subset, evaluating 24 configurations across four key parameters: $\tau_{\text{base}} \in \{0.25, 0.30\}$, $\varepsilon_{\text{spatial}} \in \{1.0, 1.5, 2.0\}$, $\tau_{\text{tile}} \in \{0.15, 0.20\}$, $\varepsilon_{\text{semantic}} \in \{0.30, 0.40\}$. Secondary parameters were fixed to default values.

Building on Stage-A, a focused random search \,[27] (Stage-B) was performed, sampling approximately 18 configurations over secondary parameters within the promising region identified earlier. The distribution of results is presented in Fig.\,\ref{fig:pr_tradeoff}, where each point corresponds to a candidate configuration in the Precision--Recall plane, with its F1-score encoded by both size and color. The scatter plot clearly illustrates the Pareto frontier: Stage-A candidates cluster toward high recall but low precision, whereas Stage-B samples extend the frontier outward, yielding more balanced trade-offs. Notably, configuration B18 (Precision = 0.595, Recall = 0.778, F1 = 0.670, Time = 0.107 s) demonstrates substantial recall recovery while maintaining acceptable precision. Complementary sensitivity analysis in Fig.\,\ref{fig:param_sens} further confirms that base and tile \emph{confidence thresholds} are the most influential factors, whereas \texttt{nms\_iou} and \texttt{confidence\_boost\_factor} play stabilizing roles. The ranked results in Table~\ref{tab:top_configs} corroborate these observations: coarse grid search is effective for identifying recall-oriented regimes, while focused random search refines the precision--recall balance and uncovers practical deployment candidates such as B18 and B14.

\begin{figure}[!t]
\centering
\includegraphics[width=\linewidth]{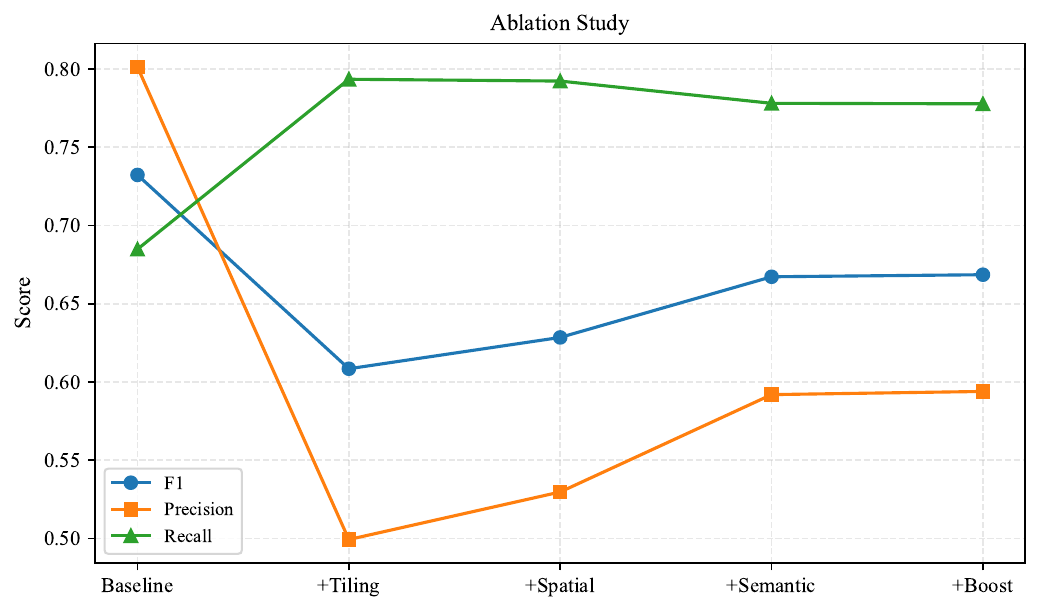}
\caption{Ablation study trajectory. The blue line corresponds to precision, the orange line to recall, and the green line to F1-score. The curves illustrate the sequential effects of adding tiling, spatial gating, semantic gating, and quality-aware reweighting, highlighting the ``recall-first, then precision recovery'' process.}
\label{fig:ablation_curve}
\end{figure}
\begin{figure}[!t]
\centering
\includegraphics[width=\linewidth,keepaspectratio]{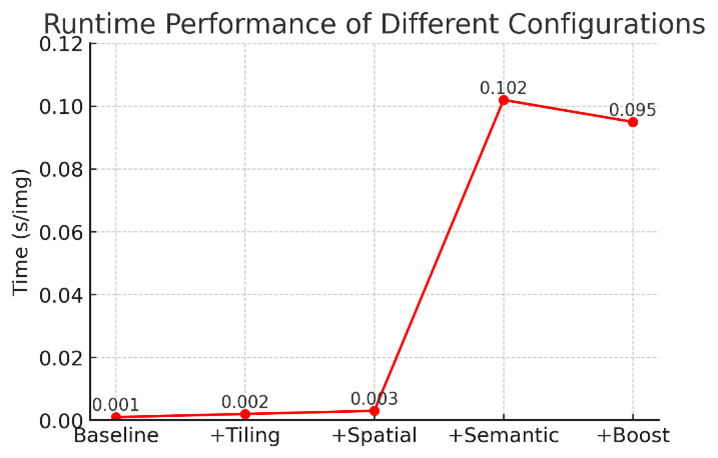}
\caption{Runtime performance of different configurations (line graph). Baseline, tiling, and spatial gating incur negligible overhead ($<$0.005 s), while semantic gating dominates runtime. Quality-aware reweighting slightly reduces latency compared to semantic gating alone.}
\label{fig:runtime_line}
\end{figure}

\begin{figure}[!t]
\centering
\includegraphics[width=\linewidth,keepaspectratio]{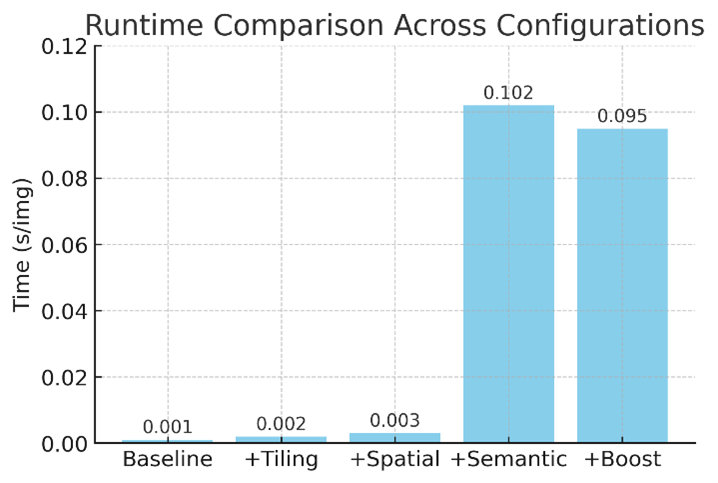}
\caption{Runtime comparison across configurations (bar chart). Semantic gating introduces the largest increase in latency (0.102 s per image), whereas other modules remain lightweight. Reweighting marginally decreases runtime by accelerating the elimination of low-quality candidates during NMS.}
\label{fig:runtime_bar}
\end{figure}
\subsection{Ablation Study}\label{sec:ablation}
The experiment further quantifies the marginal contribution of each proposed component by conducting a progressive ablation experiment, where modules are incrementally enabled in the sequence Baseline $\to$ Tiling $\to$ Spatial Gating $\to$ Semantic Gating $\to$ Quality-aware Reweighting. The aggregated results are summarized in Table~\ref{tab:ablation}, and the performance trajectory is plotted in Fig.\,\ref{fig:ablation_curve}.

The baseline YOLO detector yields the highest raw precision (0.801) but is recall-limited (0.685), leading to an overall F1 of 0.732. This reflects the well-known shortcoming of single-stage detectors in dense UAV imagery, where low-confidence small objects are discarded prematurely. Adding tiling drastically increases recall to 0.793 by reintroducing low-confidence detections from overlapping windows, but at the expense of precision (0.499) due to large numbers of redundant or spurious candidates, which lowers F1 to 0.608. The inclusion of spatial gating partially corrects this imbalance: by enforcing geometric consistency across overlapping detections, it removes isolated false positives while retaining clustered candidates, thereby raising precision to 0.530 with recall largely preserved (0.792), improving F1 to 0.628. The most decisive gain comes from semantic gating, which evaluates feature-space consistency within spatial clusters. This step filters out heterogeneous groups (e.g., visually dissimilar false positives), allowing precision to climb to 0.592 while still maintaining strong recall (0.778). As a result, the F1-score reaches 0.667---a nearly 6-point increase over the tiling-only setting. However, this improvement comes with a runtime cost (0.102 s/image), as semantic gating requires ResNet-18 feature extraction and DBSCAN clustering. Finally, quality-aware reweighting calibrates the scores of validated clusters, modestly increasing F1 to 0.669 and slightly reducing runtime to 0.095 s because higher-confidence candidates are pruned earlier during class-balanced NMS.

Together, these results outline a clear progression: tiling is indispensable for recall recovery, spatial gating restores geometric reliability, semantic gating enforces visual consistency, and reweighting provides lightweight calibration. The trajectory plotted in Fig.\,\ref{fig:ablation_curve} visually reinforces this narrative, showing a ``recall-first, then precision-recovery'' evolution that mirrors the design principle of detect broadly, then verify selectively.

\begin{figure*}[t]
    \centering
    \includegraphics[width=\textwidth,keepaspectratio]{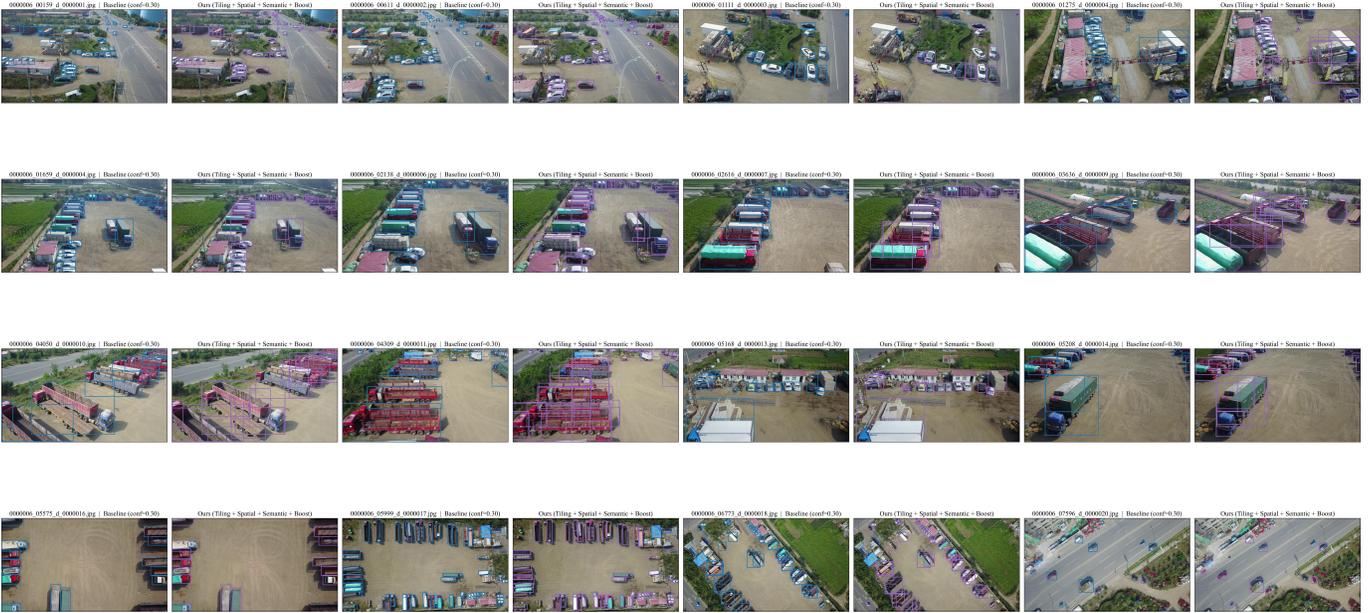}
    \caption{Qualitative comparisons across multiple VisDrone scenes. For each sample, the left image shows the baseline YOLO (conf = 0.30), and the right shows the full pipeline (Tiling + Spatial Gate + Semantic Gate + Reweighting). The method recovers many missed small/distant objects along dense rows and at long range. In ordinary scenes, near duplicate boxes may appear around tile boundaries; this behavior can be mitigated with tile-aware deduplication.}
    \label{fig:qualitative_gallery}
\end{figure*}

\subsection{Runtime Analysis}\label{sec:runtime}
In addition to accuracy, the experiment analyzed runtime performance to assess computational efficiency across different configurations. The results are summarized in Table~\ref{tab:top_configs} and Table~\ref{tab:ablation}, and visualized in Fig.\,\ref{fig:runtime_line} and Fig.\,\ref{fig:runtime_bar}. Unless otherwise stated, all latency numbers in this section refer to the post-processing stage only and exclude the base detector’s forward pass. The baseline YOLO detector achieved near-instantaneous inference, requiring only \textbf{0.001 s} per image, highlighting its suitability for strict real-time applications. Introducing tiling and spatial gating produced only marginal overhead, increasing runtime to 0.002 s and 0.003 s per image, respectively. This confirms that both modules are computationally lightweight.

By contrast, semantic gating introduced the most significant cost, raising average latency to 0.102 s due to ROI feature extraction and feature-space clustering with ResNet-18. This nearly two-orders-of-magnitude increase reflects the trade-off between improved robustness and efficiency. Interestingly, the final step of quality-aware reweighting slightly reduced runtime to 0.095 s, since higher confidence scores facilitated faster elimination of low-quality candidates during the final CB\mbox{-}NMS stage. As illustrated in Fig.\,\ref{fig:runtime_line}, runtime remains nearly flat for the first three configurations and rises sharply with semantic gating. The bar chart in Fig.\,\ref{fig:runtime_bar} further emphasizes the gap: while baseline and lightweight modules operate below 0.005 s, semantic gating dominates the runtime budget. Nonetheless, even with semantic gating enabled, the system maintains an inference time below 0.1 s per image, which remains feasible for near real-time UAV detection where recall is prioritized.

\subsection{Qualitative Analysis}\label{sec:qualitative}
Fig.\,\ref{fig:qualitative_gallery} presents side-by-side comparisons across a variety of VisDrone scenes, where each sample juxtaposes the baseline detector and the full pipeline (tiling + spatial + semantic + reweighting). In small-object--dense scenes (e.g., rows of parked vehicles, truck depots, and far-field traffic), the method consistently recovers many missed instances along long object rows and in distant perspective regions. These cases highlight the intended effect of the pipeline: tiling exposes low-confidence micro-objects; spatial gating suppresses isolated outliers; and semantic gating preserves visually consistent clusters, turning a large set of weak tile proposals into coherent detections. Qualitatively, this translates into denser, more complete object coverage in congested areas and at long range, aligning with the recall gains measured quantitatively (Table~\ref{tab:ablation}).

In ordinary or lightly crowded scenes containing a few medium/large objects, the method occasionally produces duplicate boxes around the same target. This behavior is expected when overlapping tiles generate near-identical proposals that (i) form small spatial clusters and (ii) exhibit homogeneous appearance, allowing some duplicates to survive the semantic gate and class-balanced NMS. Practically, these duplicates are localized near tile boundaries and large object extents; they have limited impact on recall but can depress precision in such scenes---consistent with the precision dip seen after enabling tiling in the ablation (Table~\ref{tab:ablation}, Fig.\,\ref{fig:ablation_curve}). Overall, the qualitative evidence confirms the central claim: the proposed spatial--semantic gating is particularly valuable in small-object-dominant scenarios, recovering numerous true instances that the baseline misses; in ordinary scenes, a lightweight data deduplication step (e.g., tile-aware deduplication suggested in Fig.\,\ref{fig:qualitative_gallery}) could potentially restore the precision lost to tiling without sacrificing the recall gains that make the method attractive for UAV detection.

\section{Conclusion}
The paper introduced a recall-first, validate-later post-processing pipeline that turns tiling redundancy into group evidence via a spatial--semantic dual gate and quality-aware reweighting, followed by class-balanced NMS. On VisDrone2019-DET, the method improves recall by $+0.093$ (from $0.685$ to $0.778$) with $\mathrm{F1}{=}0.669$ and an average post-processing latency of $0.095$ s/image. While the precision reduction reflects an inherent single-frame trade-off, the pipeline is appealing for recall-sensitive applications. Limitations include the cost of semantic gating and reliance on clustered object distributions. Future work will incorporate temporal cues, explore lightweight backbones/knowledge distillation to reduce latency, and integrate it into end-to-end trainable architectures.

\end{document}